%% file: main.tex
\PassOptionsToPackage{table,dvipsnames}{xcolor}
\documentclass[]{style}

\usepackage{microtype}
\usepackage{amsfonts}
\usepackage{amsmath}
\usepackage{amsthm}
\usepackage{xcolor}
\usepackage{graphicx}     
\usepackage{booktabs}     
\usepackage{tabularx}
\usepackage{longtable}
\usepackage{xltabular}
\usepackage{makecell}
\usepackage{multirow}
\usepackage{wrapfig}
\usepackage{float}
\usepackage{subcaption}
\usepackage{caption}
\usepackage{enumitem}     
\usepackage{pifont}       
\usepackage{url}
\usepackage[most]{tcolorbox} 
\usepackage{fontawesome5}    
\usepackage{algpseudocode}
\usepackage[linesnumbered,lined,boxed,commentsnumbered,ruled,longend]{algorithm2e}
\usepackage[T1]{fontenc}
\usepackage{xspace}       

\usepackage{tikz}
\usepackage[edges]{forest}
\usetikzlibrary{
  arrows.meta,
  positioning,
  calc,
  shapes.symbols,
  shapes.geometric,
  shapes.misc
}

\theoremstyle{plain}

\theoremstyle{definition}

\theoremstyle{remark}

\definecolor{themecolor}{HTML}{37D2A6}        
\definecolor{themecolor_light}{HTML}{9BE9D3}
\definecolor{themecolor_lighter}{HTML}{CDF4E9}

\newcommand{\ghlink}[1]{\faIcon{github}\,\href{#1}{GitHub}}
\newcommand{\weblink}[1]{\faIcon{globe}\,\href{#1}{Website}}
\newcommand{\disco}{\textsc{DisCo}\xspace}
\newcommand{\sysname}{\textsc{AREX-Skill}\xspace}

\colorlet{bandcolor}{CornflowerBlue!15}      

\title{Repo-To-Skill: Distilling GitHub Repositories Into AI4AI Skills} 

\author{Jianlyu Chen$^{1,2\dagger\ddagger}$, Yuyang Hu$^{1,3\dagger\ddagger}$, Hongjin Qian$^{1\dagger}$, Jiawei Liu$^{2\dagger}$, Wenqing Wei$^{1,2\dagger\ddagger}$, Xiaolong Chen$^{2}$ \\ Defu Lian$^{2\ast}$, Zhicheng Dou$^{3\ast}$, Chaozhuo Li$^{1}$, Qiwei Ye$^{1}$, Zheng Liu$^{1,4\ast}$}

\contribution[\dagger]{Equal Contribution}
\contribution[\ddagger]{Work done during an internship at BAAI}
\contribution[\ast]{Corresponding authors}

\affiliation{$^{1}$Beijing Academy of Artificial Intelligence, $^{2}$University of Science and Technology of China, \\ $^{3}$Renmin University of China, $^{4}$Hong Kong Polytechnic University
}

\input{sections/abstract}

\metadata[\faEnvelope\ Correspondence]{liandefu@ustc.edu.cn, ~dou@ruc.edu.cn, ~zhengliu1026@gmail.com} 
\metadata[{\raisebox{-0.2ex}{\includegraphics[height=1em]{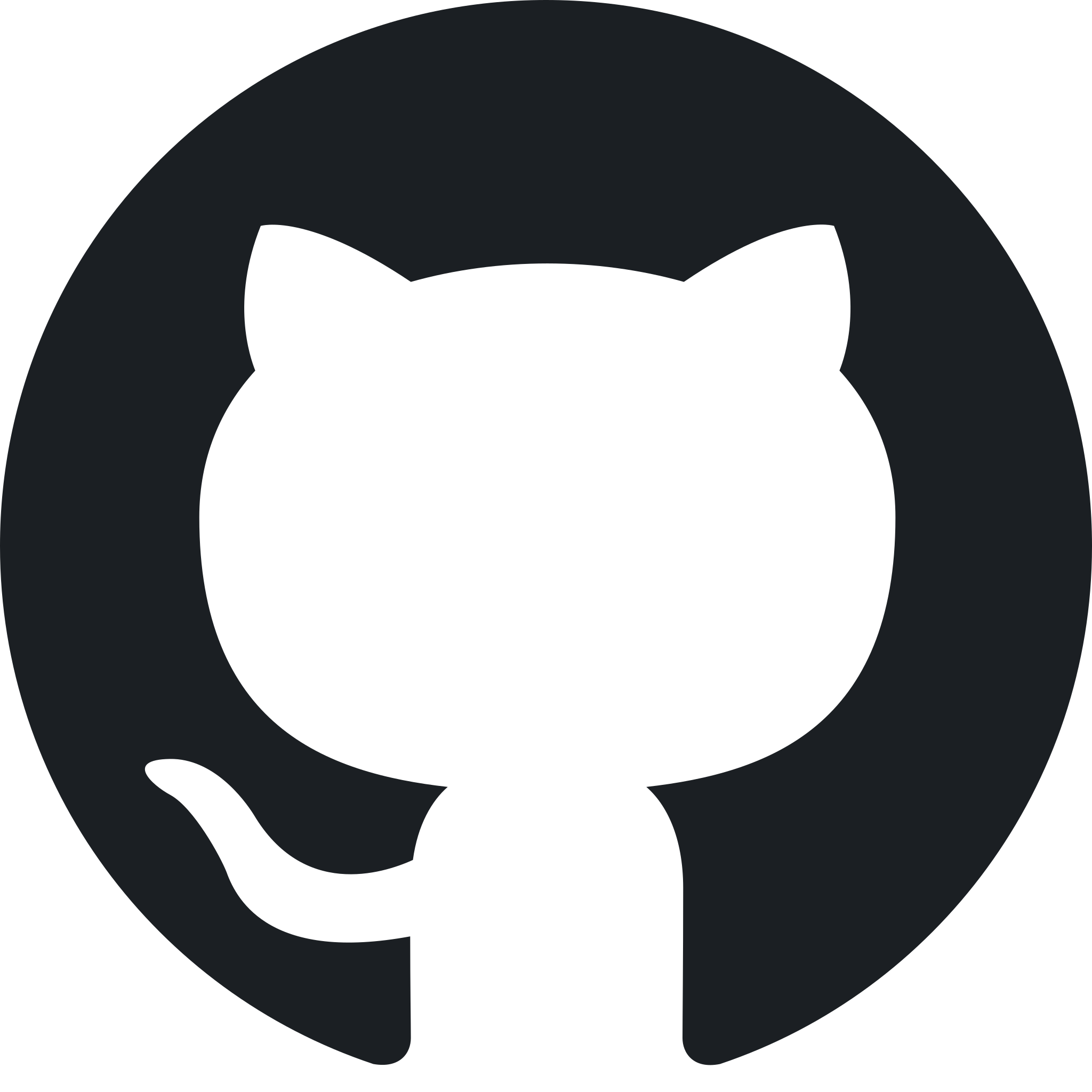}}\ Code}]{\url{https://github.com/VectorSpaceLab/AREX-Skill}}

\begin{document}

\maketitle

\begin{figure*}[h]
    \centering
    \includegraphics[width=\textwidth]{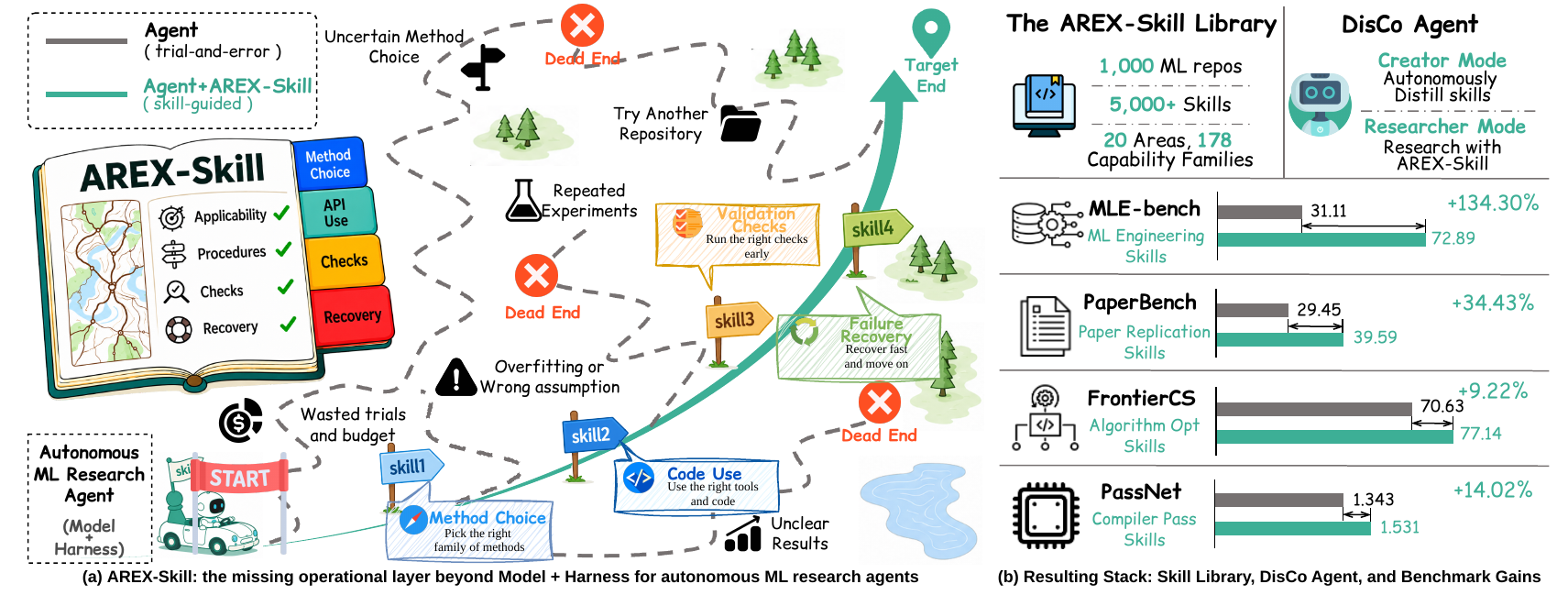}
    \caption{
    \textbf{The impact of reusable skills on autonomous research agents.}
    (a) AREX-Skill adds the missing operational layer beyond the model and harness, replacing trial-and-error exploration with reusable procedures that support the task's operational steps.
    (b) The skill library and DisCo agent form the resulting stack, and skill augmentation improves performance across four benchmarks.
    }
    \label{fig:skill_overview}
\end{figure*}

\input{sections/introduction}
\input{sections/method}
\input{sections/disco}
\input{sections/library}
\input{sections/experiments}
\input{sections/related}
\input{sections/conclusion}

\bibliographystyle{assets/plainnat}
\bibliography{citation}
\clearpage

\appendix

\input{appendix/implementations}

\input{appendix/repository-coverage}

\end{document}

%% file: sections/abstract.tex
\abstract{
Autonomous agents are beginning to carry out machine-learning (ML) research end to end. These agents combine a model backbone with a harness for planning, execution, memory, and verification, but this architecture still leaves domain-specific know-how outside the agent. We call this missing layer \emph{operational knowledge}, the know-how that separates knowing a method from making it work. That knowledge is not absent from the field. It appears in repositories and papers, but in forms written for human readers and too large to load during a task. Once distilled into compact, verified \emph{skills}, this knowledge can be reused across tasks rather than rediscovered during each run.

We present \textbf{\disco}, a skill-powered research agent that creates skills and uses them during research. Its distillation runs in two complementary forms: task-agnostic, condensing the field's widely used repositories into reusable skills, and task-oriented, producing the skills a concrete task calls for. The former, applied across the open ecosystem, yields the \textbf{\sysname Library}, with 5,000+ verified skills distilled from 1,000 widely used ML repositories and organized into 20 areas and 178 capability families. With the GPT-5.5 backbone, research harness, and downstream execution budget held fixed, the skill-equipped research agent scores 134.3\% higher on \textbf{MLE-bench}, 34.4\% higher on \textbf{PaperBench}, 9.2\% higher on \textbf{FrontierCS}, and 14.0\% higher on \textbf{PassNet} than the same agent without skills. These gains come from adding distilled operating context under that fixed setup.
}

%% file: sections/introduction.tex
\section{Introduction}
\label{sec:intro}

Autonomous agents are beginning to execute larger parts of the machine-learning (ML) research pipeline, from implementing methods to running experiments and comparing results~\citep{lu2024aiscientist,yamada2025aiscientistv2,schmidgall2025agentlaboratory}. ML research is a natural testbed because much of its practice unfolds in software, where coding agents have proved most capable~\citep{jin2026arbor,dong2026longhorizon}.

Like any agentic system, these research agents rest on two modules: a \emph{model} that supplies understanding, reasoning, planning, and execution, and a \emph{harness} that supplies orchestration, memory, verification, and iterative refinement. The model improves with frontier generations, and the harness improves through engineering practice~\citep{karpathy2026autoresearch}. ML research, however, is expertise-intensive, which means success depends on knowing which methods and tools to use, when to use them, and how to use them correctly. Neither component carries this expertise. The model's prior is broad but fixed, while the harness controls procedure but does not supply domain content. We call the missing layer \emph{operational knowledge}. \textbf{Operational knowledge is what separates knowing a method from making it work: the expertise that binds the field's methods and tools to the task at hand.} In ML research, it ranges from choosing appropriate methods and experimental settings to using package APIs correctly, configuring training pipelines, and handling common implementation and evaluation pitfalls. This knowledge exists in abundance, scattered through repositories and papers yet organized for no task in particular.

Without operational knowledge, an agent loses budget within a task and fails to reuse what it infers across tasks. Within a task, it must infer package behavior through trial and error, and mistakes surface only after budgets have been spent on misconfigured runs. Across tasks, those discoveries are not retained as reusable context. The knowledge must reach the agent in a form it can command: discoverable, loadable, and ready to use. \emph{Skills} offer this form~\citep{anthropic2025agentskills}. A skill packages one piece of know-how. A \texttt{SKILL.md} file states what the skill is for, when it applies, and how to proceed. Reference documents carry evidence, and scripts automate routine actions. Because a skill opens with a summary and unfolds only on demand, an agent can hold thousands of skills yet read just the few a task needs, allowing the knowledge layer to scale without crowding the context. The harness still governs how the agent researches, while skills determine what it knows when research begins. Figure~\ref{fig:skill_overview}(a) illustrates this missing layer: beyond the model and harness, skills turn unguided trial and error into guided execution.

The skill representation addresses how operational knowledge is consumed, but not how the skills are obtained. The relevant source material already exists in repositories and papers, but it is written for human readers and is too large to load during a task. We distill this material into compact, operational, and verified skills that fit the agent's context budget. The difficulty lies in the sources, which drift with every release, omit many practical pitfalls, and state methods without the know-how needed to make them work. The methodological problem is to produce operational knowledge automatically and at scale from declarative sources.

We meet this challenge with \disco, a skill-powered research agent that both creates skills and researches with them. \disco leaves the model and the harness unchanged and builds the operational-knowledge layer through skill distillation in two complementary forms. \emph{Task-agnostic distillation} works ahead of time, condensing the field's widely used repositories and everyday tools into reusable skills that any research task can draw on. \emph{Task-oriented distillation} works on demand. Given a concrete task, \disco explores the knowledge the task touches and produces the skills it calls for. Under either form, no skill enters the layer without verification. Each candidate is checked, repaired where possible, and recorded with any remaining gaps.

Task-agnostic distillation, run across the open ecosystem, yields the \sysname Library, whose current repository snapshot contains 5,000+ skills distilled from 1,000 widely used ML repositories and organized by a router over 20 areas and 178 capability families. Each repository is distilled into a skill graph, and the library-level router narrows a request to the relevant graphs. The paper also studies paper-derived and task-oriented skills constructed for the evaluations.

To isolate the effect of distilled skills, we compare the same research agent with and without them on \textbf{MLE-bench}~\citep{chan2025mlebench}, \textbf{PaperBench}~\citep{starace2025paperbench}, \textbf{FrontierCS}~\citep{mang2025frontiercs}, and \textbf{PassNet}~\citep{liu2026passnet}, holding the GPT-5.5 backbone, research harness, and downstream execution budget fixed. Skill construction is completed before downstream execution begins, so skills are the only variable at run time. Under these matched settings, the skill-equipped agent scores 134.3\% higher on MLE-bench, 34.4\% higher on PaperBench, 9.2\% higher on FrontierCS, and 14.0\% higher on PassNet. Figure~\ref{fig:skill_overview}(b) summarizes the resulting stack and benchmark gains.

\noindent\textbf{Our contributions are summarized as follows:}
\begin{itemize}[leftmargin=*,nosep]
\item[\ding{182}] We identify \emph{operational knowledge} as the missing layer of autonomous research agents, complementing the model and the harness. The harness governs how an agent researches, while operational knowledge determines what it knows when research begins.
\item[\ding{183}] We present \disco, a skill-powered research agent that both creates skills and researches with them. Its skill distillation runs in two complementary forms, task-agnostic and task-oriented, and admits no skill without verification.
\item[\ding{184}] We build the \sysname Library by scaling \disco across the open ecosystem, yielding 5,000+ verified skills distilled from 1,000 widely used ML repositories, organized into 20 areas and 178 capability families, and exposed through a library-level router.
\item[\ding{185}] We evaluate distilled skills on \textbf{MLE-bench}, \textbf{PaperBench}, \textbf{FrontierCS}, and \textbf{PassNet} under a matched backbone, harness, and downstream execution budget, and observe consistent gains on all four, up to 134.3\% on MLE-bench.
\end{itemize}

%% file: sections/method.tex
\section{Operational Knowledge for Autonomous Research}
\label{sec:method}

We first formalize an autonomous research task and then isolate the knowledge layer left unspecified by the standard view of a model and a harness. Section~\ref{sec:method:preliminary} defines the task and agentic system. Section~\ref{sec:method:operational_knowledge} defines \emph{operational knowledge}. Section~\ref{sec:disco} instantiates this layer with \disco.

\subsection{Preliminary}
\label{sec:method:preliminary}

A research task can be formalized as
\begin{equation}
    \tau=(q,\mathcal{D},\mathcal{E},g),
    \label{eq:research-task}
\end{equation}
where $q$ states the problem, $\mathcal{D}$ is the data and material given with it, $\mathcal{E}$ is the environment in which the work is carried out, including the tools it exposes and the budget it bounds, and $g$ is the target the outcome must meet. To solve $\tau$ is to produce a set of \emph{artifacts} $y$, including code, models, experimental results, and reports, that fulfill $g$ in $\mathcal{E}$. This view treats research as the mapping $\tau\mapsto y$ from a problem and its surrounding material to artifacts that satisfy $g$.

An \emph{agentic system} carries out this mapping on its own. It is conventionally described by two components,
\begin{equation}
    \mathcal{A}=(M_\theta,H),
    \label{eq:agentic-system}
\end{equation}
the LLM backbone $M_\theta$, which supplies understanding, reasoning, planning, and execution, and the harness $H$, which supplies orchestration, memory, verification, and iterative refinement. Together, they turn the mapping into a loop of reasoning, action, and observation that runs until $g$ is met or the budget is spent. At step $t$, with history $h_t=(a_1,o_1,\ldots,a_{t-1},o_{t-1})$, the agent acts and the environment responds:
\begin{equation}
    a_t \sim \pi_\theta(a\mid \tau,h_t,H),
    \qquad
    o_t=\mathcal{E}(a_t),
    \label{eq:agent-policy}
\end{equation}
and the artifacts $y$ are what the trajectory leaves behind. Nearly all progress on autonomous research agents has come from strengthening these two components. Backbones grow stronger with each frontier generation, while harness engineering continues to mature~\citep{li2025fmagent,chen2026toward,jin2026arbor}. For research tasks, however, the two-component view leaves the agent's domain-specific operational knowledge unspecified.

\subsection{Defining Operational Knowledge}
\label{sec:method:operational_knowledge}

An agent asked to improve a model on an unfamiliar dataset must decide which method suits the problem, which package implements it, how that package expects the data to be laid out, which configurations are appropriate, and which pitfalls can invalidate an otherwise plausible run. The agent can infer these choices through trial and error by proposing a plan, running it, inspecting the failure, and revising. Such failures consume the same budget used for evaluation, and a misconfigured run may spend a large share of $\mathcal{E}$ before producing a meaningful measurement. What the agent needs at the moment of decision is not an answer it could eventually reach, but one that is already executable. Neither component of $\mathcal{A}$ supplies this knowledge. The backbone's prior is broad but fixed, while the harness controls procedure but does not supply domain content. We call what is missing \emph{operational knowledge}, and write a research agent as
\begin{equation}
    \mathcal{A}_{\mathrm{res}}=(M_\theta,H,\mathcal{K}),
    \label{eq:research-agent}
\end{equation}
where $\mathcal{K}$ is the operational knowledge made available to the agent as explicit operating context, so that Eq.~\eqref{eq:agent-policy} becomes $a_t \sim \pi_\theta(a\mid \tau,h_t,H,\mathcal{K})$.

\textbf{Operational knowledge is what turns knowing about a domain into being able to act in it. It binds the field's methods and tools to the problem at hand by specifying \emph{what} can solve it, \emph{when} each candidate applies, and \emph{how} it should be used.} It has two constituents. The first turns \emph{knowledge into capability}: methods, code, models, and APIs are packaged into units the agent can actually invoke. The second turns \emph{capability into usage policy}: every unit carries the conditions, reasons, and procedures that govern its use. The two constituents are complementary. Capability without policy gives the agent tools without selection criteria. Policy without capability gives it advice without an executable interface. This separation also distinguishes $\mathcal{K}$ from $H$. The harness specializes \emph{how} the agent explores, while $\mathcal{K}$ specializes \emph{what} the agent knows to consider.

Declarative sources state facts about methods, APIs, or design choices. A paper reports that a technique improves accuracy. A repository documents what an API accepts. A blog post explains why a trick works. All of this is useful, but none of it directly specifies a course of action for a given problem. Declarative knowledge states what holds, while operational knowledge translates those facts into task-level actions. The latter must be derived from the former.

In current practice, this derivation is manual. An expert reads the papers, repositories, and technical blogs that contain the declarative material, wraps the useful parts into custom tools and scripts, and writes the skills and usage instructions that tell an agent when and how to invoke them. The result can be genuine operational knowledge, but its cost scales with expert labor and with the domain, stack, and release for which it was written. The declarative material it draws on, meanwhile, is abundant and continually updated. \textbf{The central methodological problem is to produce operational knowledge automatically and at scale from declarative sources.}

%% file: sections/disco.tex
\section{\texorpdfstring{\disco}{DisCo}: Producing and Using Operational Knowledge}
\label{sec:disco}

\disco instantiates the operational-knowledge layer as skills, constructs them through skill distillation, and uses them as operating context during research. Section~\ref{sec:method:skills} defines skills and skill graphs. Section~\ref{sec:method:ml_knowledge_distillation} describes the distillation mechanism in task-agnostic and task-oriented forms. Section~\ref{sec:method:skill_usage} brings production and use together in \disco.

\begin{figure*}[t]
    \centering
    \includegraphics[width=\textwidth]{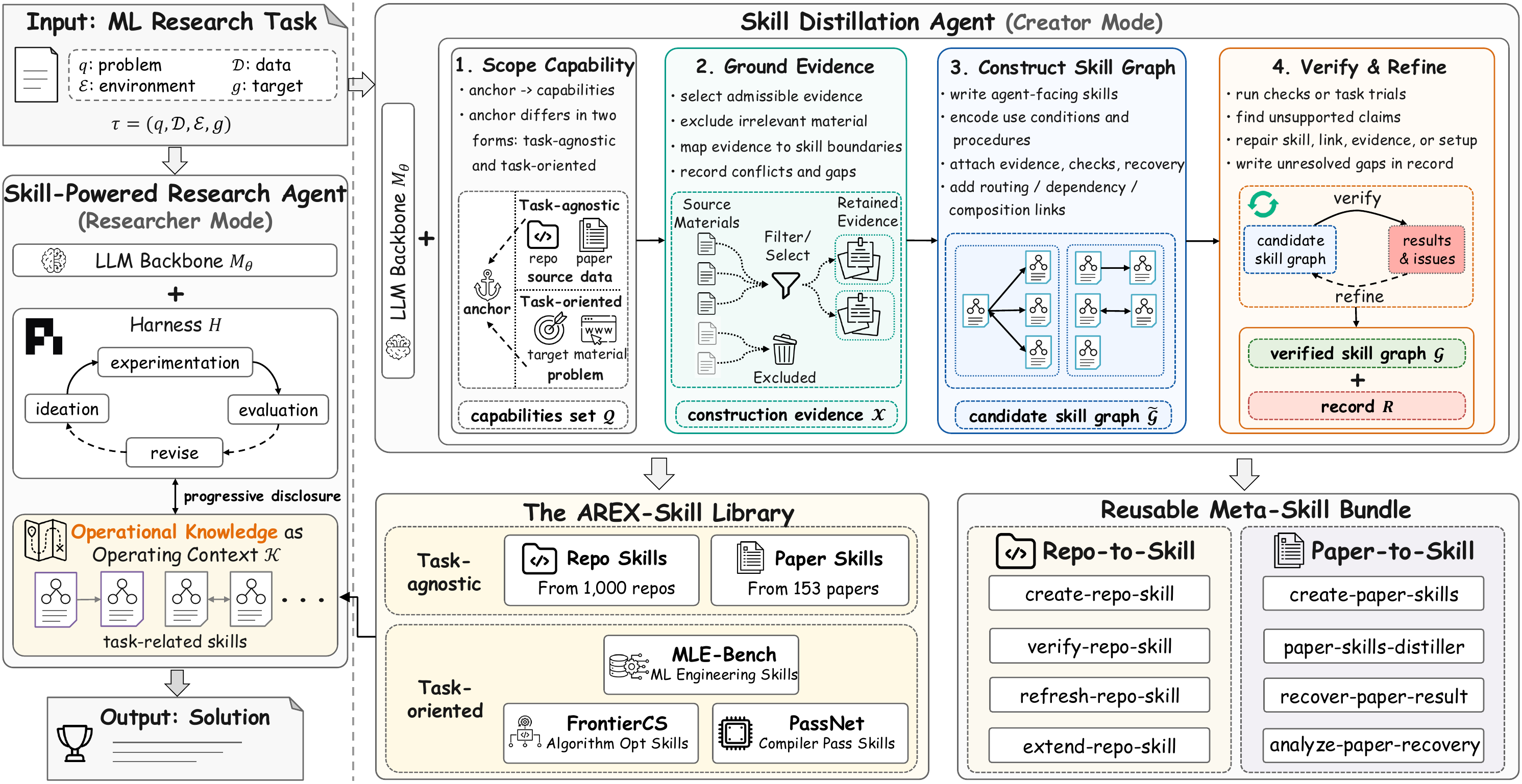}
    \caption{\textbf{\disco produces and uses operational knowledge.} A research task $\tau=(q,\mathcal{D},\mathcal{E},g)$ is solved by an agent with backbone $M_\theta$, harness $H$, and operating context $\mathcal{K}$. In creator mode, an anchor $z$, either a source $c$ or a task $\tau$, is scoped into capabilities $\mathcal{Q}$, grounded in evidence $\mathcal{X}$, packaged as a candidate graph $\tilde{\mathcal{G}}$, and verified into an accepted graph $\mathcal{G}$ with construction record $R$. In researcher mode, the agent loads only the relevant branch of accepted graphs from the \sysname Library and uses those skills as $\mathcal{K}$ during execution.}
    \label{fig:method}
\end{figure*}

\subsection{Skills and Skill Graphs}
\label{sec:method:skills}

We instantiate $\mathcal{K}$ as a set of \emph{skills}~\citep{anthropic2025agentskills},
\begin{equation}
    \mathcal{K}=\{S_1,\ldots,S_m\},
    \label{eq:skill-set}
\end{equation}
where $\mathcal{K}$ is the set of skills the agent holds at a given moment, and the problem of producing operational knowledge becomes the problem of producing skills. Skills are a practical carrier for this layer. A skill is self-contained, agent-facing, and already supported by modern agentic systems such as Claude Code~\citep{anthropic2025claudecode} and Codex~\citep{openai2025codexcli}. Making one available to an agent requires no change to $M_\theta$ or $H$. The skill becomes part of the operating context the agent may draw on, allowing operational knowledge to move across compatible harnesses and accumulate across tasks.

In \sysname, a skill is organized in three layers,
\begin{equation}
    S=(\underbrace{\texttt{SKILL.md}}_{\text{knowledge interface}},\ \underbrace{\texttt{references/}}_{\text{knowledge substrate}},\ \underbrace{\texttt{scripts/}}_{\text{execution interface}}),
    \label{eq:skill-structure}
\end{equation}
each serving a different purpose. \texttt{SKILL.md} is the \emph{knowledge interface}. As the only layer read up front, it states what the agent must know to use the skill and outlines the rest. It serves as the entry point, carries the standard operating procedure, and routes onward to deeper material or sibling skills. Its content provides the information an agent needs before loading deeper material, including goals, key concepts, tool usage, pointers, worked examples, and known failure modes. \texttt{references/} is the \emph{knowledge substrate}, the deeper material that \texttt{SKILL.md} points to and that is loaded only when needed, following the principle of progressive disclosure~\citep{anthropic2025agentskills}. It holds API documentation, algorithmic detail, parameter configurations, and related material. \texttt{scripts/} is the \emph{execution interface}, consisting of executable wrappers with defined inputs and outputs that the agent invokes rather than reimplements. The three layers directly realize the two constituents of Section~\ref{sec:method:operational_knowledge}. \texttt{scripts/} and \texttt{references/} turn knowledge into capability, while \texttt{SKILL.md} turns capability into usage policy and keeps the cost of holding a skill low enough for an agent to hold thousands of them.

A single source often contains more operational knowledge than one skill should hold, so \sysname organizes the skills distilled from one source as a \emph{skill graph}
\begin{equation}
    \mathcal{G}=(\mathcal{S},\mathcal{L}),\quad
    \mathcal{S}=\{S_i\}_{i=1}^{n},\ n\geq 1,\quad
    \mathcal{L}\subseteq
    \{(S_i,S_j)\in\mathcal{S}\times\mathcal{S}\mid i\neq j\}.
    \label{eq:skill-graph}
\end{equation}
The graph contains an entry skill that states the source scope and routes to component skills for package functions, method stages, or protocol elements. Each link $(S_i,S_j)\in\mathcal{L}$ encodes a routing, dependency, or composition relation, and $\mathcal{L}$ may be empty when a source yields a single skill or several independent ones. Progressive disclosure operates over this graph. The agent reads the entry point, follows the links its problem calls for, and leaves the rest unopened, so $\mathcal{K}$ at any moment contains only the part of the graph needed by the task.

\subsection{Skill Distillation}
\label{sec:method:ml_knowledge_distillation}

What remains is to produce such graphs automatically. We call this \emph{skill distillation}, the process of reworking declarative source knowledge into operational knowledge that directly supports task solving. Every run, regardless of what triggers it, follows the same four-stage process. Writing $z$ for the \emph{anchor} that initiates the run and $\mathcal{C}$ for the declarative source material it can reach, whether held in advance or searched for,
\begin{equation}
    z
    \ \xrightarrow{\ \mathsf{scope}\ }\
    \mathcal{Q}
    \ \xrightarrow{\ \mathsf{ground}\ }\
    \mathcal{X}
    \ \xrightarrow{\ \mathsf{construct}\ }\
    \tilde{\mathcal{G}}
    \ \xrightarrow{\ \mathsf{verify}\ }\
    (\mathcal{G},R),
    \label{eq:distillation}
\end{equation}
where $\mathcal{Q}$ is the set of capabilities the run decides to cover, $\mathcal{X}\subseteq\mathcal{C}$ is the evidence gathered to support them, $\tilde{\mathcal{G}}$ is the candidate skill graph assembled from that evidence in the three layers of Eq.~\eqref{eq:skill-structure}, and the accepted graph $\mathcal{G}$ comes with a construction record $R$ that retains the evidence used, the checks performed, and any unresolved gaps. The four stages answer four questions in turn: which capabilities matter, what supports them, how they become skills, and whether those skills hold. The two forms of distillation differ in the anchor, and that choice determines downstream source selection and the verification signal.

\paragraph{Task-agnostic distillation.} Here the anchor is a source, $z=c\in\mathcal{C}$, such as a repository, a paper, or a tutorial. The run asks what the source makes possible and packages the answer into long-lived skills that are built ahead of time and available to any task that later needs them. Scoping consists of \emph{Source Understanding} followed by \emph{Capability Identification}: first establish what the artifact is and how it is organized, then decide which of its capabilities are worth exposing. Grounding is \emph{Knowledge Extraction}, which gathers the evidence supporting each capability from the source itself. Construction consists of \emph{Tool Encapsulation} and \emph{Skill Packaging}: executable parts are wrapped behind stable interfaces, and the three layers are then assembled into a connected graph. Verification is \emph{Skill Verification}, performed before anything is admitted. Distilling repositories such as \texttt{sentence-transformers}, \texttt{AlphaFold}, and \texttt{vLLM}, or papers that introduce reusable methods and techniques, yields skills that can be reused across tasks.

\paragraph{Task-oriented distillation.} Here the anchor is a problem, $z=\tau$, and the source material is not given in advance but actively sought. The run asks what solving the task demands and produces the skills that a problem of this kind requires. Scoping consists of \emph{Task Decomposition} followed by \emph{Capability Gap Analysis}: the task is broken into the capabilities it calls for, after which the capabilities the agent cannot already supply are isolated. Grounding is \emph{Source Discovery}, which searches for material covering those gaps, so that $\mathcal{X}$ is assembled rather than selected. Construction is \emph{Skill Generation}, which distills that material into skills for the task. Verification closes the run as before. Optimizing an open-ended algorithmic problem or entering a Kaggle competition are tasks of this kind. The skills are produced on demand but remain reusable for the class of problems they address.

Whichever anchor initiates the run, verification is what separates distillation from summarization. No skill is admitted on the strength of its sources alone, and any gap that survives the checks is recorded in $R$ rather than hidden.

\subsection{Creator and Researcher Modes}
\label{sec:method:skill_usage}

The two halves of the framework, a layer that must be produced and a layer that must be used, meet in a single agent. We define \disco as a research agent that both creates skills and researches with them, operating in two modes over the same backbone and harness. In \emph{creator mode}, \disco carries out the distillation of Section~\ref{sec:method:ml_knowledge_distillation}. It scopes, grounds, constructs, and verifies, then deposits the accepted graph $\mathcal{G}$ in the \sysname Library (Section~\ref{sec:library}). In \emph{researcher mode}, \disco is the agent of Eq.~\eqref{eq:research-agent}, solving a task $\tau$ with $\mathcal{K}$ drawn from that library. The library connects the two modes, with creator mode writing accepted graphs and researcher mode retrieving them. Figure~\ref{fig:method} gives a compact overview.

The modes are deliberately asymmetric in cost. Creator mode is paid once per source and amortized over every task that later draws on the result. Researcher mode pays only for what a task actually opens. This cost asymmetry makes the layer scalable in the settings we study. Distillation runs offline at whatever breadth the source ecosystem allows, while the agent solving $\tau$ inherits the result without re-deriving it.

Within researcher mode, the remaining question is how a constructed graph is consumed. \disco follows the progressive disclosure principle~\citep{anthropic2025agentskills}. Rather than reading the full graph, the agent first sees a router or graph entry skill that summarizes candidate skill graphs by scope and intended use. It then chooses an entry point and opens only the skills needed during research execution. Each generated skill begins with a use description that helps the agent recognize its relevance and load the skill's procedures, evidence, checks, and recovery actions into $\mathcal{K}$.

The links in $\mathcal{G}$ extend the same selection process beyond the first skill. For instance, an entry skill can route to component skills for setup, evaluation, diagnosis, or repair. These links make routing, dependency, and composition relations visible to the agent. During a research task, the agent can move from an opened skill to a referenced skill when needed without reading unrelated parts of the graph. In this way, skill graphs provide selective operating context while the harness continues to control planning, action selection, tool use, and observation.

The interface adds operating context rather than a new control loop. A skill graph is not a new way of running an agent, which keeps the missing knowledge layer separate from harness design. The same distilled skills can serve any compatible harness that exposes agent-readable skills, while each harness retains its own planning policy, tool interface, and execution loop.

%% file: sections/library.tex
\section{The \texorpdfstring{\sysname}{AREX-Skill} Library}
\label{sec:library}

The \sysname Library is the persistent operational-knowledge store used by \disco. Creator mode writes accepted skill graphs into the library, and researcher mode retrieves a task-relevant branch as operating context. We organize the library by source anchor. The public repository snapshot covers 1,000 ML repositories, while paper-derived and task-oriented skill graphs form separate collections in the library.

\begin{figure*}[t]
    \centering
    \includegraphics[width=\textwidth]{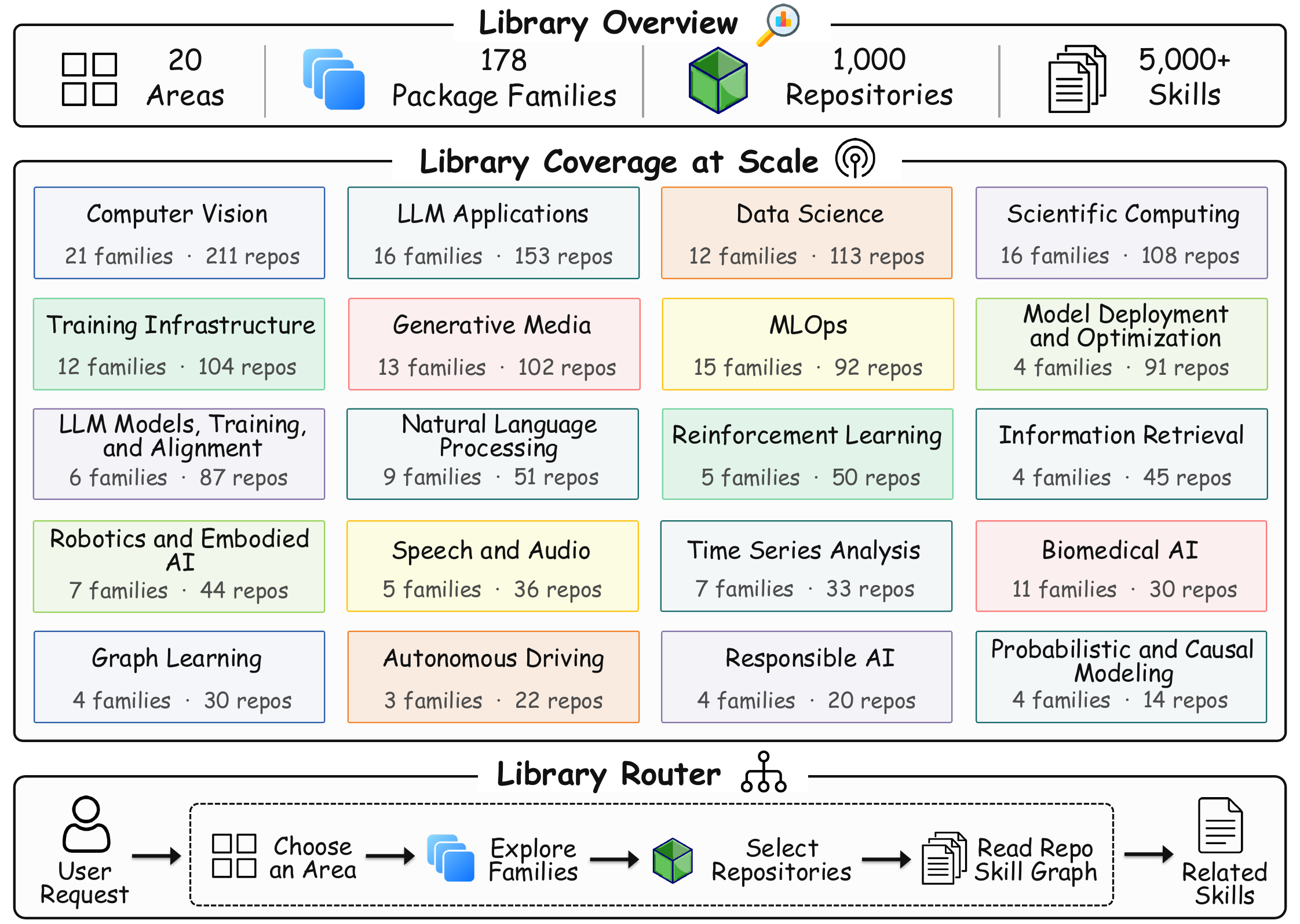}
    \caption{\textbf{Repository collection and router in the \sysname Library.} The collection contains 5,000+ skills distilled from 1,000 ML repositories and indexed by 20 areas and 178 capability families. A repository may appear under multiple area-to-family paths when it supports multiple capabilities, so area and family memberships are overlapping rather than disjoint. During research, the router narrows a request from area to family to repository graph, allowing the agent to load only the relevant skills.}
    \label{fig:library-overview}
\end{figure*}

Repository graphs are organized with a two-level capability taxonomy and a generated router. The taxonomy groups repositories by area and family, while the router exposes these levels before a repository graph is opened. Figure~\ref{fig:library-overview} summarizes the collection and retrieval path.

\subsection{Repository Collection}
\label{sec:library:repositories}

\paragraph{Scope.} We select 1,000 ML repositories based on open-source visibility and practical use, with GitHub stars among the curation signals. The collection spans model implementations, training and deployment systems, data and evaluation tools, and scientific software. It is a curated snapshot of commonly used ML software rather than an exhaustive partition of the ecosystem.

\paragraph{Skill Graph construction.} For each repository, \disco constructs a verified skill graph using the task-agnostic procedure in \Cref{sec:method:ml_knowledge_distillation}. Construction uses GPT-5.5 and GPT-5.6-sol with \texttt{xhigh} reasoning effort, at an average allocation of about \$40 per repository. The evidence boundary includes repository source, documentation, examples, tests, scripts, and configuration. The graph decomposes supported workflows into skills for data preparation, training, inference, evaluation, serving, troubleshooting, or maintenance. References and scripts retain the details needed to execute these workflows.

Before inclusion, \disco checks each graph's content and usability using assertion-backed cases and safe repository-native examples, tests, CLI checks, tiny-fixture checks, or smoke scripts when available. A failure attributed to the graph triggers local repair and reruns of the relevant checks. Appendix~\ref{app:repo-skill-construction} gives the construction and verification details.

\paragraph{Coverage.} The repository snapshot contains 5,353 skills across 1,000 repository graphs, organized into 20 areas and 178 capability families. It records 2,209 exact assignments of repositories to area-to-family paths, with 700 repositories appearing in more than one family. Appendix~\ref{app:repo-coverage} lists the routed coverage. Paper-derived and task-oriented graphs are not included in these counts.

\subsection{Taxonomy and Routing}
\label{sec:library:routing}

\paragraph{Taxonomy construction.} The repository taxonomy is fixed before final assignment. We freeze a short repository summary for each source and induce a two-level tree of areas and families from these descriptions. Stars, URLs, and pre-existing category fields are excluded from the model input. The LLM-assisted pipeline proposes a complete tree, evaluates it on 100 stable batches of 10 repositories with separate locator and judge calls, and revises it using judge reviews together with deterministic fit and family-load statistics. Provisional placements are used only for diagnosis. Final routes are assigned after the taxonomy is fixed. Appendix~\ref{app:repo-skill-construction} gives the implementation details.

\paragraph{Repository assignment.} After the taxonomy is fixed, each verified graph is classified against exact area-to-family paths. The original repository serves as the primary evidence source, while the generated graph is used for navigation. Each assignment requires a rationale, repository evidence, and confidence. Keyword-only, dependency-only, optional-integration, and example-only matches are rejected. A repository may receive multiple assignments, and a graph remains unclassified when no exact family is supported.

\paragraph{Router generation and use.} Accepted assignments are compiled into a router from repository and assignment indexes, which are used to generate the area and family pages. During research, the model starts with the router description, follows the relevant area-to-family path, opens the selected repository graph, and loads only the skills, references, or scripts needed for the current step. It can open several graphs when they provide distinct capabilities, but does not force a route when no family is a close fit. This implements progressive disclosure~\citep{anthropic2025agentskills}, so only the selected branch enters the operating context $\mathcal{K}$.

\subsection{Paper-Derived and Task-Oriented Skills}
\label{sec:library:beyond-repositories}

\paragraph{Paper-derived skills.} Papers provide a second task-agnostic source anchor. \disco decomposes each paper into module-level skills covering method components, data or evaluation procedures, and implementation workflows. Each module is checked in isolation, followed by a bounded recovery experiment that excludes the original implementation repository.

For the current study, we apply this workflow to prior papers selected for the 20 PaperBench targets. This produces 636 paper-derived skills from 153 source papers. Related-work context determines which source-paper skills are available to each target, while the target paper and its released artifacts are excluded as skill sources. The resulting skills remain reusable beyond the target. Appendix~\ref{app:paperbench-skill-construction} gives the construction protocol, and Section~\ref{sec:exp-paper} evaluates the resulting pools. The same workflow can be extended to additional papers as verification budget permits.

\paragraph{Task-oriented skills.} We also construct skills for benchmarks whose operational knowledge is defined by a task interface and feedback signal. MLE-bench uses one descriptive graph for each of 75 competitions, built through task decomposition, source discovery, and bounded diagnostic trials, with competition-specific content excluded. FrontierCS uses one recovery-oriented graph shared across its 188 Agent Track tasks. PassNet uses one benchmark-level graph for FX-graph inspection, pattern matching, semantics-preserving rewrites, Triton implementation, and performance diagnosis. Appendix~\ref{app:mle-skill-construction}, Appendix~\ref{app:frontier-skill-construction}, and Appendix~\ref{app:passnet} describe these constructions. Sections~\ref{sec:exp-mle}, \ref{sec:exp-alg}, and \ref{sec:exp-kernel} evaluate them.

%% file: sections/experiments.tex
\section{Experiments}
\label{sec:experiments}

We evaluate \disco beyond the public repository snapshot by constructing paper-derived skill pools for PaperBench and task-oriented skill graphs for MLE-bench, FrontierCS, and PassNet. In each setting, \disco constructs skills under the task's source constraints, evaluation protocol, and verification conditions, and we measure whether the resulting operating context improves a fixed research agent.

\subsection{Setup}

Distilled skills act as operating context rather than as a new control loop, so they can be attached to a fixed research harness. We use Codex~\citep{openai2025codexcli} as the harness and keep the \textbf{GPT-5.5} backbone with \texttt{xhigh} reasoning effort fixed across conditions. The only controlled factor is whether the agent is equipped with \disco-distilled skills (\emph{with skills}) or not (\emph{without skills}).

\paragraph{MLE-bench.} On MLE-bench~\citep{chan2025mlebench}, we evaluate the \emph{full} suite of \textbf{75} competitions across all three difficulty tiers (Low, Medium, High). We report MLE-bench's headline Any-Medal score by difficulty split. Final grading uses the benchmark's held-out grader. For each task, we construct a dedicated operational-knowledge skill graph before attempting it, distilling knowledge sources collected through web search while excluding the original competition webpage and competition-specific content. Skill construction and benchmark execution use separate per-task budgets. The with- and without-skills conditions are compared under a \emph{matched running budget}, while the one-time construction budget is separate and is not counted in either run-time condition. The full procedure is described in Appendix~\ref{app:mle-skill-construction}.

\paragraph{PaperBench.} On PaperBench~\citep{starace2025paperbench}, we run the full task set of \textbf{20} papers, evaluated with the benchmark's official replication grader. For each paper, before attempting the reproduction, we select relevant works cited in the target paper's related-work section and distill skills from these papers and their corresponding repositories, while excluding the target paper itself and any accompanying released code or artifacts. As in MLE-bench, skill construction and benchmark execution use separate per-task budgets. The with- and without-skills conditions are compared under a \emph{matched running budget}, while the one-time construction budget is separate and is not counted in either run-time condition. The full procedure is described in Appendix~\ref{app:paperbench-skill-construction}.

\paragraph{FrontierCS.} FrontierCS~\citep{mang2025frontiercs} comprises open-ended computer-science problems whose solution quality is objectively measurable even when the optimum is unknown. We evaluate all \textbf{188} Agent Track tasks through Harbor~\citep{Harbor_Framework}. Each task run has a 5-hour budget, and the agent container is limited to 2 CPUs and 4\,GiB of RAM. The agent inspects the problem, implements a C++ solver, and may improve it through iterative submissions. The retained task score is the best score from any intermediate submission. The test cases for each task inherit that task's own judge limits, which range from 0.25 to 100\,s and from 128\,MiB to 2\,GiB across the suite. Following the leaderboard convention, we report average score together with mean per-task usage in steps, tool calls, and tokens. For this benchmark, we construct a single skill graph shared across all tasks. Construction and refinement are completed before evaluation, after which the graph is frozen for all skill-equipped runs. Further details are provided in Appendix~\ref{app:frontier-skill-construction}.

\paragraph{PassNet.} PassNet~\citep{liu2026passnet} targets graph-compiler pass generation, where each task requires constructing a pattern matcher and rewriter that preserve semantics while improving execution performance. We report four complementary metrics following the benchmark convention. AS Score is the primary benchmark score. For each sample, PassNet computes ES($t$) as the geometric mean of rectified subgraph speedups over tolerance levels $t \in [-10, 4]$, then aggregates these per-sample scores using the official weighted geometric-mean protocol before averaging across samples. The rectified speedup assigns a floor value of 0.1 to incorrect or unmatched passes. The other three metrics are computed at tolerance $t=-5$. G-Mean Speedup is the geometric mean of end-to-end speedups over numerically correct subgraphs. Correctness is the fraction of subgraphs whose outputs match the eager-mode reference at the specified tolerance. {Fast\_1} is the share of correct subgraphs that are at least as fast as eager execution. All evaluations are conducted on NVIDIA A100-SXM4-40GB. The full procedure is described in Appendix~\ref{app:passnet}.

\subsection{Main Results: MLE-bench (Full)}
\label{sec:exp-mle}

\Cref{tab:mle} compares Codex with and without distilled skills against strong agents from the public MLE-bench leaderboard. All reported results use the full 75-task suite.

\begin{table*}[t]
\centering
\caption{\textbf{Main results on MLE-bench (full, 75 competitions).} We report
Any Medal (\%) across the Low, Medium, High, and full 75-task splits. Public
baselines are copied from the official MLE-bench leaderboard
(\url{https://github.com/openai/mle-bench}) and report mean$\pm$SEM over runs.
\sysname values follow the same convention over three repeated runs.
\textbf{Bold} marks the best score in each column.}
\label{tab:mle}
\small
\setlength{\tabcolsep}{4pt}
\renewcommand{\arraystretch}{1.1}
\begin{tabular}{@{}llcccc@{}}
\toprule
\textbf{Agent} & \textbf{Backbone} &
\makecell[c]{\textbf{Low}\\\textbf{(n=22)}} &
\makecell[c]{\textbf{Medium}\\\textbf{(n=38)}} &
\makecell[c]{\textbf{High}\\\textbf{(n=15)}} &
\makecell[c]{\textbf{All}\\\textbf{(n=75)}} \\
\midrule
Famou-Agent 2.0~\citep{li2025fmagent} & Gemini-3-Pro-Preview & 80.30$\pm$1.52 & 64.04$\pm$2.32 & 42.22$\pm$2.22 & 64.44$\pm$1.18 \\
AIBuildAI~\citep{zhang2026aibuildai} & Claude-Opus-4.6 & 77.27$\pm$0.00 & 61.40$\pm$0.88 & 46.67$\pm$0.00 & 63.11$\pm$0.44 \\
CAIR MARS+~\citep{chen2026mars} & Gemini-3-Pro-Preview & 78.79$\pm$1.52 & 60.53$\pm$1.52 & 44.44$\pm$2.22 & 62.67$\pm$0.77 \\
MLEvolve~\citep{du2026mlevolve} & Gemini-3-Pro-Preview & 80.30$\pm$1.52 & 57.89$\pm$1.52 & 42.22$\pm$2.22 & 61.33$\pm$1.33 \\
PiEvolve~\citep{sai2025pievolve} & Gemini-3-Pro-Preview & 80.30$\pm$1.52 & 58.77$\pm$0.88 & 40.00$\pm$0.00 & 61.33$\pm$0.77 \\
Thesis~\citep{thesis} & GPT-5 & 65.15$\pm$1.52 & 45.61$\pm$7.18 & 31.11$\pm$2.22 & 48.44$\pm$3.64 \\
R\&D-Agent~\citep{zhang2026reasoning} & GPT-5 & 68.18$\pm$2.62 & 21.05$\pm$1.52 & 22.22$\pm$2.22 & 35.11$\pm$0.44 \\
\midrule
Codex & GPT-5.5 & 42.42$\pm$6.60 & 31.58$\pm$1.52 & 13.33$\pm$3.85 & 31.11$\pm$2.22 \\
Codex + \sysname & GPT-5.5 & \textbf{86.36$\pm$2.62} & \textbf{69.30$\pm$3.16} & \textbf{62.22$\pm$2.22} & \textbf{72.89$\pm$1.18} \\
\bottomrule
\end{tabular}
\end{table*}
\FloatBarrier

\paragraph{Skills improve Codex on MLE-bench.} Adding skills raises the overall Any-Medal score from 31.11\% to 72.89\%, a gain of 41.78 percentage points and a 134.3\% relative improvement, without changing the agent backbone. The gains are consistent across all difficulty tiers: 43.94 points on Low, 37.72 points on Medium, and 48.89 points on High. The relative gain is particularly pronounced on High tasks, where the score rises from 13.33\% to 62.22\%, corresponding to a 366.8\% improvement, or 4.67 times the no-skill score. Under the same agent and task budget, this result supports our central hypothesis that distilled operational knowledge can improve ML research performance.

\paragraph{The gain does not require a new harness.} Codex with skills also surpasses the strongest public baseline in \Cref{tab:mle}, improving the overall score from 64.44\% to 72.89\% (+8.45 points). The corresponding gains over the best public score in each difficulty tier are 6.06 points on Low, 5.26 points on Medium, and 15.55 points on High. This comparison uses vanilla Codex with added distilled skills, without a custom execution harness, specialized agent orchestration strategy, or modified control loop. The comparison isolates the contribution of externalized operational knowledge under the fixed Codex harness.

\paragraph{The advantage grows with task difficulty.} The largest gains appear on High-difficulty tasks, both against Codex without skills (+48.89 points) and against the strongest public baseline (+15.55 points). Harder tasks typically expose the agent to a larger space of libraries, implementations, and optimization choices, making unguided trial and error more costly. The observed trend is consistent with skills guiding the agent's search toward applicable tools, validated workflows, and explicit checks. Instead of spending its budget exploring unsuitable code paths, the agent can enter a productive region of the solution space earlier and focus its iterations on implementation and optimization choices that directly affect the target metric. This stronger trend on difficult tasks suggests that the value of distilled skills may increase as ML problems become more complex.

\begin{table*}[t]
\centering
\caption{\textbf{Comparison between vanilla GPT-5.5 Codex and Codex equipped with
distilled skills on PaperBench (20 papers).} $\Delta$ denotes the improvement
of Codex+\sysname over vanilla Codex. \textbf{Bold} indicates the better score
between the two variants.}
\label{tab:paperbench-skill}
\resizebox{\textwidth}{!}{%
\small
\setlength{\tabcolsep}{5pt}
\renewcommand{\arraystretch}{1.05}
\begin{tabular}{@{}l|l|cc|c@{}}
\toprule
\textbf{Paper} & \textbf{ICML Topic}
& \textbf{GPT-5.5 Codex}
& \textbf{GPT-5.5 Codex+\sysname}
& \textbf{$\Delta$} \\
\midrule

adaptive-pruning
& Deep Learning: LLMs
& 33.42
& \textbf{38.05}
& \textcolor{red}{+4.63}
\\

all-in-one
& Probabilistic Methods
& 52.93
& \textbf{54.70}
& \textcolor{red}{+1.77}
\\

bam
& Probabilistic Methods - Variational Inference
& 56.65
& \textbf{58.83}
& \textcolor{red}{+2.18}
\\

bbox
& Deep Learning: LLMs
& 17.30
& \textbf{28.49}
& \textcolor{red}{+11.19}
\\

bridging-data-gaps
& Theory: Domain Adapt. \& Transfer Learning
& 14.16
& \textbf{31.42}
& \textcolor{red}{+17.26}
\\

fre
& Deep RL
& 14.06
& \textbf{23.88}
& \textcolor{red}{+9.82}
\\

ftrl
& Reinforcement Learning: Deep RL
& 1.50
& \textbf{17.17}
& \textcolor{red}{+15.67}
\\

lbcs
& Data-Centric AI
& 26.60
& \textbf{29.10}
& \textcolor{red}{+2.50}
\\

lca-on-the-line
& Deep Learning: Robustness
& 22.08
& \textbf{39.59}
& \textcolor{red}{+17.51}
\\

mechanistic-understanding
& Deep Learning: LLMs
& 45.67
& \textbf{47.85}
& \textcolor{red}{+2.18}
\\

pinn
& Deep Learning
& 40.64
& \textbf{58.10}
& \textcolor{red}{+17.46}
\\

rice
& Deep RL
& 7.94
& \textbf{48.51}
& \textcolor{red}{+40.57}
\\

robust-clip
& Deep Learning: Robustness
& 30.14
& \textbf{32.35}
& \textcolor{red}{+2.21}
\\

sample-specific-masks
& Misc. Aspects of ML: General ML Techniques
& \textbf{57.11}
& 52.04
& \textcolor{blue}{-5.07}
\\

sapg
& Deep RL
& 18.15
& \textbf{35.06}
& \textcolor{red}{+16.91}
\\

sequential-neural
& Probabilistic Methods
& 41.67
& \textbf{65.37}
& \textcolor{red}{+23.70}
\\

stay-on-topic
& Deep Learning: LLMs
& \textbf{32.31}
& 27.79
& \textcolor{blue}{-4.52}
\\

stochastic-interpolants
& Generative Models
& 41.12
& \textbf{42.28}
& \textcolor{red}{+1.16}
\\

test-time-model-adaptation
& Distributions Shift and OOD
& 26.28
& \textbf{30.77}
& \textcolor{red}{+4.49}
\\

what-will-my-model-forget
& Deep Learning: Everything Else
& 9.35
& \textbf{30.45}
& \textcolor{red}{+21.10}
\\

\midrule

\textbf{Average Score}
&
&
29.45
&
\textbf{39.59}
&
\textcolor{red}{+10.14}
\\

\bottomrule
\end{tabular}
}
\end{table*}

\subsection{Main Results: PaperBench (Full)}
\label{sec:exp-paper}

\Cref{tab:paperbench-skill} reports per-task replication scores. Distilled skills raise the average replication score from 29.45\% to 39.59\% (+10.14 points, a 34.4\% relative improvement), with the largest gains on rice (+40.57), sequential-neural (+23.70), and what-will-my-model-forget (+21.10). Red values report the gain of \sysname over the no-skill baseline.

\paragraph{Skills improve Codex on PaperBench.} Adding skills raises the average replication score from 29.45\% to 39.59\%, a gain of 10.14 points and a 34.4\% relative improvement, without changing the agent backbone. The effect is broad rather than concentrated in a few outliers. Skills improve the score on 18 of the 20 tasks and degrade it on only 2, matching the pattern observed on MLE-bench and FrontierCS, where skills help across most tasks rather than only on a small subset.

\paragraph{The largest gains occur on low-baseline tasks.} The improvement is markedly larger in relative terms on tasks where Codex without skills starts from a low replication score. On ftrl, the score rises from 1.50 to 17.17, an 11.4$\times$ increase. On rice, it rises from 7.94 to 48.51, a 6.1$\times$ increase. On what-will-my-model-forget, it rises from 9.35 to 30.45, a 3.3$\times$ increase. In contrast, tasks where the no-skill baseline is already moderate to high, such as all-in-one (52.93) and bam (56.65), see smaller absolute gains (+1.77 and +2.18, respectively). This pattern matches the FrontierCS recovery analysis, where skills help most when unguided attempts stall on implementation details, environment setup, or method-specific components.

\paragraph{A small number of tasks regress under skills.} Two tasks score lower with skills than without: sample-specific-masks (57.11 $\to$ 52.04, $-$5.07) and stay-on-topic (32.31 $\to$ 27.79, $-$4.52). Both have no-skill scores above the 20-task average (29.45), suggesting a possible retrieval-precision failure. Retrieved skill content may occasionally distract from a task-specific strategy that the base agent would otherwise discover on its own. This is consistent with a modest precision-recall trade-off in skill retrieval. For papers whose replication depends on a narrow, idiosyncratic implementation choice not well covered by the constructed skill graph, following the retrieved skill may pull the agent away from an approach it would have converged on unaided. Better routing or an explicit fallback to unguided reasoning when retrieved skills are a poor match may reduce this failure mode.

\subsection{Main Results: FrontierCS (Agent Track)}
\label{sec:exp-alg}

\Cref{tab:frontier-cs-agent} compares the two controlled Codex conditions with representative entries from the public FrontierCS Agent Track leaderboard. Both Codex conditions cover the full 188-task suite and differ only in access to the frozen skill graph.

\begin{table*}[t]
\centering
\caption{\textbf{Main results on the FrontierCS Agent Track (188 tasks).}
We report aggregate Score and mean trajectory usage per task. Values for the
public reference systems are taken directly from the official FrontierCS
leaderboard (\url{https://frontier-cs.org/\#leaderboard}), whereas the two
Codex rows report results from our controlled runs. All entries use the standard
five-hour task budget. \textbf{Bold} marks the highest score.}
\label{tab:frontier-cs-agent}
\small
\setlength{\tabcolsep}{6pt}
\renewcommand{\arraystretch}{1.1}
\begin{tabular}{@{}llrrrr@{}}
\toprule
\textbf{Agent} & \textbf{Backbone} & \textbf{Score} & \textbf{Avg. Steps} & \textbf{Avg. Tool Calls} & \textbf{Avg. Tokens} \\
\midrule
Claude Code & Claude Opus 4.8  & 74.5\phantom{0}          & 355.4 & 145.7 & 14.72M \\
Claude Code & Qwen3.7 Max      & 61.9\phantom{0}          & 133.9 & 139.1 & 13.85M \\
Gemini CLI & Gemini 3.1 Pro    & 60.2\phantom{0}          & 74.3  & 41.6  & 2.00M  \\
\midrule
Codex      & GPT-5.5           & 70.63 & 55.9 & 64.7 & 2.46M \\
Codex + \sysname  & GPT-5.5     & \textbf{77.14} & 88.7 & 105.0 & 4.47M \\
\bottomrule
\end{tabular}
\end{table*}
\FloatBarrier

\paragraph{Skills improve FrontierCS scores.} Providing Codex with the distilled operating context increases its Score from 70.63 to 77.14, an absolute gain of 6.51 points and a relative improvement of 9.22\%. At the task level, skills improve performance on 74 tasks and leave 66 effectively unchanged. Improved tasks gain 22.23 points on average, whereas degraded tasks lose 8.76 points on average. The aggregate positive change is 3.91$\times$ the magnitude of the aggregate negative change. A paired bootstrap over all 188 tasks yields a 95\% confidence interval of $[3.41, 9.83]$ points for the mean improvement.

\paragraph{Skills recover lower-scoring problems.} Stratifying by the no-skill score, the largest lift occurs on the 47 tasks below 50, whose mean rises from 19.43 to 45.99 (+26.56), with 30 improved tasks crossing the 50-point threshold. More broadly, the skills condition shifts the score distribution upward, increasing the number of tasks that reach moderate, high, and near-complete scores. These results suggest that operating guidance is particularly valuable when unguided exploration fails to find a workable formulation or implementation.

\paragraph{Skills improve performance beyond raw resource scaling.} Codex accesses at least one skill file on 180 of 188 tasks. On the 102 tasks for which the skills run does not invoke a sub-agent, the mean paired gain remains 5.54 points. On the 86 tasks that use sub-agents, the gain is 7.66 points. Skills also lead to more active search overall, increasing tokens, steps, and tool calls. However, per-task gains are essentially uncorrelated with the additional usage. Spearman's $\rho$ is 0.006 for tokens, 0.014 for steps, and 0.015 for tool calls. Additional usage alone does not account for the score improvement.

\paragraph{A stronger score-efficiency frontier.} Even after aggregating sub-agent usage, Codex + \sysname uses 4.47M tokens per task. The Claude Code configurations with Claude Opus 4.8 and Qwen3.7 Max use 14.72M and 13.85M, respectively, which are 3.29$\times$ and 3.10$\times$ as many tokens, while scoring 2.64 and 15.24 points lower. Codex + \sysname also uses 24.5--27.9\% fewer tool calls and 33.8--75.1\% fewer steps than these two entries. Under the leaderboard's reported accounting, Codex + \sysname Pareto-dominates both Claude Code configurations across Score, tokens, steps, and tool calls.

\subsection{Main Results: PassNet}
\label{sec:exp-kernel}
PassNet evaluates agents on graph-compiler pass generation, where a solution must construct a pattern matcher and rewriter that preserve graph semantics while improving execution performance. We keep the Codex harness and GPT-5.5 backbone fixed and vary only whether the agent receives the \disco-distilled PassNet skill, allowing us to isolate the effect of the operating context. The results are shown in Table~\ref{tab:passnet-main}.

\begin{table}[t]
\centering
\small
\setlength{\tabcolsep}{4pt}
\caption{\textbf{Main results on PassNet eval list (200 samples).} The Codex rows use the same GPT-5.5 backbone
and harness. The only controlled difference is access to the distilled
PassNet skill. Bold marks the better Codex condition in each column.}
\label{tab:passnet-main}
\begin{tabularx}{\linewidth}{Xccccc}
\toprule
\textbf{Method} & \textbf{AS Score} & \textbf{G-Mean Speedup} & \textbf{Correctness} & \textbf{Fast\_1} & \textbf{Failed samples} \\
\midrule
Eager (reference) & 1.000 & 1.000 & 100.00\% & 100.00\% & 0 \\
TorchInductor (torch.compile) & 1.419 & 1.505 & 79.70\% & 23.60\% & 0 \\
Codex + GPT-5.5 & 1.343 & 1.5891 & 81.35\% & \textbf{28.48\%} & 14 \\
Codex + GPT-5.5 + \sysname & \textbf{1.5313} & \textbf{1.6688} & \textbf{90.76\%} & 26.72\% & \textbf{5} \\
\bottomrule
\end{tabularx}
\end{table}

\paragraph{Distilled skills improve Codex on PassNet.} Adding the PassNet skill raises AS Score from 1.343 to 1.5313, an absolute gain of 0.1883 and a 14.0\% relative improvement over the no-skill Codex baseline. The geometric-mean speedup also increases from 1.5891 to 1.6688. This improvement is consistent with skills guiding the agent toward more reliable, semantically valid, and higher-scoring passes.

\paragraph{Skills reduce invalid or incomplete agent outcomes.} The no-skill Codex baseline fails on 14 samples, whereas Codex with \sysname fails on only 5, corresponding to a 64.3\% reduction in failed samples. Correctness also rises from 81.35\% to 90.76\%, a gain of 9.41 percentage points. These results align with the role of skill graphs described in Section~\ref{sec:method:skills}. The graph provides explicit procedures, rejection criteria, and recovery actions for pass matching, correctness checking, and performance tuning.

\paragraph{Skills push Codex beyond TorchInductor on aggregate score.} TorchInductor remains a strong compiler baseline, but Codex with \sysname achieves a higher AS Score of 1.5313, compared with 1.419 for TorchInductor. This suggests that the skill-equipped agent can identify optimizations beyond those captured by the default compiler pipeline on this evaluation set.

%% file: sections/related.tex
\section{Related Work}
\label{sec:related}

\subsection{Autonomous ML Research}

Recent work studies autonomous systems for the ML research lifecycle~\citep{dong2026longhorizon}. Autonomous-discovery systems chain ideation, implementation, experimentation, and writing into end-to-end pipelines~\citep{lu2024aiscientist,yamada2025aiscientistv2,schmidgall2025agentlaboratory,tang2025airesearcher,karpathy2026autoresearch}. A complementary strand studies research \emph{ideation} in isolation. Large-scale human studies compare LLM-generated and expert research ideas~\citep{si2025novelideas}, while iterative agents generate and refine ideas over the scientific literature~\citep{baek2025researchagent}. A second cluster of work targets ML engineering and data science directly. Agents solve such tasks through propose--run--evaluate loops~\citep{jiang2025aide,yang2025rdagent}, case-based reuse of prior solutions~\citep{guo2024dsagent}, multi-agent competition pipelines~\citep{li2024autokaggle}, hierarchical task graphs~\citep{hong2024datainterpreter}, tree search over candidate pipelines~\citep{chi2024sela}, and reinforcement learning over execution feedback~\citep{liu2025mlagent}. Their progress is tracked by benchmarks that measure end-to-end ML experimentation, engineering, research, and paper reproduction~\citep{huang2024mlagentbench,chan2025mlebench,starace2025paperbench,wijk2025rebench}, alongside a related line of work on automated paper-to-code reproduction~\citep{zhou2025repro,seo2026paper2code,li2025deepcode}. Recent analyses also note how brittle agents become once they leave a single well-scoped repository~\citep{chen2026beyondswe}.

These systems treat each task largely in isolation. An agent relearns each library, configuration, and launch procedure from the underlying artifacts on every task, so this effort is not amortized across tasks. Our work is complementary to advances in agent design. Rather than introducing a new control loop, we distill a reusable operational-knowledge layer that compatible agents can consume, and we use these benchmarks to measure its effect on agent performance. Our verification further draws on adversarial multi-agent review, motivated by analyses of coordination and verification failures in multi-agent systems~\citep{cemri2026why}.

\subsection{Agent Skills}

Agent skills extend language-model agents with reusable procedural knowledge without modifying model parameters. Under the Agent Skills specification, a skill is an inspectable artifact centered on a \texttt{SKILL.md} file that describes activation conditions, procedures, and tool-use strategies, optionally accompanied by scripts and references loaded through progressive disclosure~\citep{anthropic2025agentskills}. Unlike latent policies or episodic memories, such skills are portable and explicitly modifiable artifacts. However, large-scale analysis of existing \texttt{SKILL.md} files reveals authoring inconsistencies, highlighting the difficulty of creating high-quality skills at scale~\citep{hong2026skillmd}.

Prior work has explored automatically acquiring reusable knowledge from agent experience. Voyager learns and composes executable skills across environments~\citep{wang2024voyager}, Agent Workflow Memory extracts reusable routines from past trajectories~\citep{wang2024awm}, and ExpeL distills natural-language insights from task experiences~\citep{zhao2024expel}. However, these methods mainly derive free-form code, workflows, or insights from interaction traces, making skill quality difficult to verify and failures difficult to attribute. In contrast, we distill \emph{version-specific operational knowledge} from static ML artifacts, such as papers and repositories, into \emph{provenance-grounded and verified} Agent Skills with explicit validation records.

%% file: sections/conclusion.tex
\section{Conclusion}
\label{sec:conclusion}

In this paper, we study operational knowledge as a missing layer for ML research agents. \disco fills this layer by distilling source knowledge into reusable operational-knowledge skill graphs that can be loaded as operating context while leaving the model backbone and research harness unchanged. Scaling \disco yields the \sysname Library, whose repository snapshot contains 5,000+ skills distilled from 1,000 widely used ML repositories. We also construct paper-derived and task-oriented skills for the research settings evaluated in this work.

Under a fixed GPT-5.5 Codex setup and matched downstream budgets, the skill-equipped agent scores 134.3\% higher on MLE-bench, 34.4\% higher on PaperBench, 9.2\% higher on FrontierCS, and 14.0\% higher on PassNet. These results support the central claim that autonomous research agents can improve by adding operational knowledge rather than relying only on stronger control loops. Harnesses specialize how an agent researches, while distilled skills specialize what it knows to consider when research begins.

%% file: appendix/implementations.tex
\section{\texorpdfstring{\disco}{DisCo} Implementations}
\label{app:implementations}

This appendix describes three instantiations of \disco used in this work. The public repository collection distills versioned software sources into reusable task-agnostic graphs. The PaperBench implementation distills prior papers into source-anchored module skills and assembles a target-conditioned pool for each reproduction. MLE-bench, FrontierCS, and PassNet instead anchor task-oriented construction on a competition or benchmark interface. All implementations follow the four stages in \Cref{sec:method:ml_knowledge_distillation}, but differ in source selection, graph granularity, and verification signal.

\subsection{Task-Agnostic ML Repository Skill Construction}
\label{app:repo-skill-construction}

For the task-agnostic repository collection, \disco runs the distillation procedure in \Cref{sec:method:ml_knowledge_distillation} once for each selected ML repository, following a stored repo-to-skill procedure that specifies how the four stages are carried out for this type of source. The construction unit is a self-contained operational-knowledge skill graph for one versioned upstream repository. Repository graph production uses GPT-5.5 and GPT-5.6-sol with \texttt{xhigh} reasoning effort and an average construction allocation of about \$40 per repository. The anchor $z$ is the repository snapshot together with its inspectable evidence, including metadata, source roots, documentation, examples, tests, configuration files, and repo-owned scripts. The capabilities in $\mathcal{Q}$ are those needed for package-level ML research work, such as choosing interfaces, adapting code, checking outputs, and recovering from package-specific failures. Verification combines usability cases, safe native checks, and static checks, and the graph is organized around a repository-level entry skill with focused component skills when needed.

\paragraph{Scoping and grounding fix the repository evidence boundary.} \disco first analyzes the repository before installing or executing package code. This pass builds an include/exclude evidence map over source roots, documentation, examples, tests, scripts, configuration files, and existing repo-local skills when present. It also records construction-only maps for native test or example candidates and source-script handling. Generated files, build outputs, vendored dependencies, local environments, caches, large artifacts, and unrelated development internals are excluded unless they are needed for the requested repository workflow. This boundary keeps the resulting skill graph tied to public, reusable package behavior rather than arbitrary checkout state.

After the scope is fixed, \disco prepares a private Python inspection environment. The install plan is limited to the selected evidence and user requirements, avoiding broad extras when a smaller dependency set is sufficient. Live inspection checks import names, package versions, public signatures, CLI entry points, optional backends, and small runtime behaviors. Documentation and tests establish workflow intent, while source code and live inspection confirm API and runtime claims before they enter the skill graph. Local paths and machine-specific setup remain private construction evidence.

\paragraph{Construction writes a self-contained repository graph.} The generated graph starts with a repository-level entry skill. This entry skill is kept router-like. It states when the package applies, provides minimal setup or import checks, and routes the agent to component skills for distinct workflows such as data preparation, training, evaluation, serving, troubleshooting, or repository maintenance when supported by the evidence. Each component skill covers a bounded workflow and links to nearby references or scripts for API details, command construction, data formats, validation checks, and recovery procedures. The split is determined by likely future agent tasks rather than source directory names alone.

Each repository graph is self-contained. When a future agent needs an example, helper, or validator, \disco bundles it as a reference or skill-owned script rather than requiring the agent to reopen the original repository checkout. Every graph also contains public provenance, including the source commit or equivalent identifier, package version when available, dirty state, and relative evidence paths. A compact routing metadata file records only the canonical repository identity, skill identifier, taxonomy hash, routing status, and exact assignments. The full classification rationale and supporting evidence remain in construction artifacts outside the runtime graph.

\paragraph{Verification separates runtime skills from checks.} Before a repository graph is accepted, \disco runs a verification workflow over the integrated entry skill, component skills, bundled references, and bundled scripts. The verifier creates assertion-backed usability cases, reviews content against the confirmed evidence boundary, and selects safe native examples, tests, CLI checks, tiny-fixture checks, or smoke scripts when available. Native checks are classified as \textsc{pass}, \textsc{skill\_gap}, \textsc{native\_fail}, \textsc{skip\_unsafe}, or \textsc{skip\_not\_selected}. A skill gap triggers a localized repair to the responsible skill, reference, script, or route, after which the relevant checks are rerun.

Static gates check metadata, link integrity, self-containment, provenance, routing metadata, local-path leakage, and the separation between runtime files and construction-only artifacts. Runtime files are limited to the skill graph itself, such as \texttt{SKILL.md}, \texttt{references/}, \texttt{scripts/}, and optional \texttt{sub-skills/}. Verification reports and other check-only files are written outside the runtime graph.

\paragraph{Taxonomy induction builds the routing tree before assignment.}
Repository taxonomy construction is a separate Python pipeline over short repository summary records. Input normalization accepts repository identifiers and short repository summaries, rejects duplicate repository IDs and empty summaries, stable-sorts the inventory, records SHA-256 hashes, and writes an immutable normalized JSONL file. Only the repository identifier and short repository summary are sent to the model. URLs, stars, word counts, and any pre-existing area or family fields are retained as audit metadata but excluded from model input, so the tree reflects capability semantics rather than popularity or inherited labels.

The pipeline first asks an LLM to propose a complete two-level tree of areas and families, then splits the 1,000 repositories into 100 stable batches of 10 repositories. Each batch is processed by two separate model calls. A locator provisionally assigns every repository to an exact, acceptable, poor, or no-fit area-to-family path to stress-test coverage. An independent judge then reviews whether the current tree captures the capability distinctions in that batch, identifying structural issues such as missing capabilities, overlapping families, mixed axes, ambiguous scopes, overloaded nodes, and catch-all categories. These provisional assignments are used only for diagnosis and are not published as final router assignments.

After all batches finish, an aggregator synthesizes the 100 judge reviews together with deterministic statistics. These statistics include exact, acceptable, poor, and no-fit rates, per-family provisional load, effective support for small-family review, unused and overloaded families, and outlier evidence. A reviser applies the approved synthesis by writing a complete revised tree. Before the next evaluation round begins, the workflow validates the revised schema and tree diff, including typed add, remove, rename, merge, split, and scope-rewrite effects. The default convergence gate requires at least two evaluation rounds, full batch coverage, bounded no-fit and poor-fit rates, bounded small-family and unused-family rates, no blocker issues, no major-change judge batches, no overloaded families, no unresolved revision directives, and an explicit stop recommendation from the aggregator. The run writes the final taxonomy JSON, a rendered Markdown tree, a run summary, and final outlier records. It does not write final repository assignments.

\paragraph{Classification is separate from skill generation.} After a graph passes verification, \disco classifies the original repository against the fixed taxonomy of areas and families. The repository checkout is the primary evidence source, while the generated graph is used only for navigation. Every assignment requires an assignment-specific rationale, confidence, and non-generated repository evidence. Keyword-only, dependency-only, optional-integration, and example-only matches are rejected. A repository may receive several assignments when it exposes several distinct capabilities. If no exact family is supported, the graph is recorded as unclassified rather than being forced into a weak route.

\paragraph{The library router exposes the graph through progressive disclosure.} After verification, classification, and approval, the repository graph is imported into the managed repository-skill collection, and the sibling router is rebuilt from the central repository and assignment indexes. The current public collection contains 1,000 repository graph entries under \texttt{skills/repositories/repo-skills/}, together with a model-visible \texttt{skills/repositories/repo-skills-router/}. The repository entry and component skills are agent-facing skills, but they are omitted from the initial model-visible skill list and accessed through the router. During research, researcher mode narrows the request from area to family, opens the relevant repository graph, and loads only the skills, references, or scripts needed as operating context $\mathcal{K}$. The full library is never placed into context, and the downstream agent receives only the repository operational knowledge selected for the current ML research task.

\subsection{Paper-Derived and Task-Oriented Skill Construction}

Beyond the public repository collection, we apply \disco to the four autonomous-research evaluations in Section~\ref{sec:experiments}. PaperBench uses source-anchored paper skills, with the downstream skill pool selected for each target reproduction. The other three settings use task-oriented distillation. MLE-bench constructs one graph per competition, whereas FrontierCS and PassNet each construct a single graph for the benchmark. In all four settings, skill construction is completed before downstream evaluation, and the reported with- and without-skills conditions keep the backbone, harness, and running budget fixed.

\subsubsection{MLE-bench}
\label{app:mle-skill-construction}

For MLE-bench, we run task-oriented distillation (\Cref{sec:method:ml_knowledge_distillation}) once per task and separate this one-time skill construction from downstream benchmark execution. The anchor is the competition itself, $z=\tau$. Scoping decomposes it into the capabilities required by a solution, grounding gathers task-relevant resources through web search, construction assembles a task-level skill graph, and verification uses execution feedback from diagnostic trials, all within a workspace under a bounded construction budget. During the running phase, the graph is frozen and we measure whether it helps Codex solve the task under a separate, matched execution budget. \Cref{tab:mle-phases} summarizes the two phases.

\begin{table}[t]
\centering
\caption{The two-stage MLE-bench protocol. Both limits apply per task. Skill
construction is completed before benchmark running begins.}
\label{tab:mle-phases}
\small
\begin{tabularx}{\linewidth}{l X c}
\toprule
\textbf{Phase} & \textbf{Purpose and output} & \textbf{GPU budget} \\
\midrule
Exploration & Explore useful modeling and execution decisions and finalize a
task-oriented set of descriptive skills. & $\leq 24$ GPU-hours \\
Running & Let Codex select from the finalized skill pool and optimize a full
benchmark submission. & $\leq 24$ GPU-hours \\
\bottomrule
\end{tabularx}
\end{table}

\paragraph{Scoping and grounding begin with a research plan.} The scoping and grounding stages together produce an evidence-backed plan. Each exploration iteration begins before code execution with an explicit research-planning step. Based on its initial diagnosis of the task, the model decomposes the problem and identifies the operational knowledge needed for a productive trial. The plan may ask, for example, which pretrained model is most appropriate for continued training or fine-tuning, how the data should be preprocessed, or whether a specialized domain requires consulting related work. It then guides web search and source collection, assembling the evidence $\mathcal{X}$ for the task. To prevent direct task leakage, we block the task's original competition webpage and exclude competition-specific content associated with that benchmark task. During construction, the collected evidence is synthesized into an execution-oriented skill that describes the recommended decisions, procedures, and checks. Codex reads this skill, writes the corresponding code, and executes the trial.

\paragraph{Execution feedback drives verification.} The refinement loop implements the verification stage, using execution feedback as the verification signal. After each trial, Codex inspects the execution logs and observed results, summarizes the run, and analyzes which decisions were useful or limiting. It then decides whether the current skill should be refined. When refinement is needed, the next research-plan-refine step receives exactly three forms of context:
\begin{equation}
    \mathcal{F}_{t+1}
    = \bigl(S_t,\; \operatorname{Summary}(L_t),\;
      \operatorname{Analysis}(R_t)\bigr),
    \label{eq:mle-refine-context}
\end{equation}
where $S_t$ is the previous skill, $L_t$ is the execution log, and $R_t$ is the observed result at iteration $t$. The model uses this context to revise the research plan, collect additional evidence when necessary, and produce the next skill version. The revised skill is then evaluated through another trial written and executed by Codex. This loop continues until the exploration budget is exhausted or the model determines that further refinement is unlikely to be productive.

\paragraph{Exploration optimizes learning progress, not medal attainment.} The construction phase does not require Codex to reach a medal-level score. Medal attainment may require a long, fully optimized training run, whereas the purpose of exploration is to test as many plausible modeling and implementation directions as the budget allows. We therefore favor rapid diagnostic iterations. A direction is continued when an iteration produces a useful improvement. Otherwise, the model can revise the plan or move to another direction. This criterion uses the 24 GPU-hours to accumulate broadly useful operational knowledge rather than spending most of the construction budget on completing a single submission.

\paragraph{Final skills are descriptive rather than executable.} At the end of exploration, the task's operational-knowledge skill graph is finalized as descriptive guidance only. Its skills may record task diagnoses, model and preprocessing choices, training strategies, expected observations, and checks, but contain no runnable training or inference scripts. Codex must generate and execute the task implementation from this guidance during each benchmark run. This design prevents the running phase from replaying a fixed solution script and preserves run-to-run stochasticity in agent decisions and implementation. Because each task yields a small graph of largely independent skills, the link set $\mathcal{L}$ remains light, and routing is handled by the library-level entry point rather than a deep per-task hierarchy.

\paragraph{Benchmark running uses autonomous skill selection.} In the running phase, the finalized task skill graph is made available to Codex. We do not prescribe a fixed task-to-skill mapping. Following progressive disclosure, Codex decides which skills are relevant for each task, writes the implementation, and runs the resulting solution. Unlike exploration, this phase uses its full budget to train, iterate, and optimize the benchmark target. Each task receives at most 24 GPU-hours. The no-skill condition uses the same Codex backbone and running budget but does not receive the distilled skills. The reported comparison isolates the downstream effect of access to the finalized operational knowledge under matched running budgets. The separate exploration budget is the one-time cost of constructing that knowledge.

\subsubsection{PaperBench}
\label{app:paperbench-skill-construction}

For PaperBench, the unit of reusable knowledge is a source paper rather than the downstream reproduction target. \disco distills each selected prior paper into module-level skills and then assembles a pool of these skills for a target paper. These source skills are task-agnostic with respect to future use, although the pool exposed in a PaperBench run is conditioned on the target paper's related-work context. Skill construction remains strictly separated from benchmark running.

\paragraph{Target-conditioned source collection.} For each target paper $\tau$, we analyze its related work and methodology to identify prior studies that address closely related research problems. We select up to 10 representative papers from the related-work section and collect their open-source repositories when available. The target paper itself and its released code or artifacts are excluded as skill sources. The selected prior papers provide the methodological specification, while their repositories may provide implementation evidence for data processing, model components, training, inference, and evaluation.

\paragraph{Source-anchored module skills.} Each selected source paper is decomposed into a small set of functionally bounded modules rather than summarized as a single skill. A module skill states the reusable capability, its inputs and outputs, applicable conditions, dependencies, configuration, procedure, and validation contract. Supporting scripts or references are included when needed to make the module executable or self-contained. The boundaries follow the paper's reusable method and experimental components rather than the directory structure of an accompanying implementation. In the current protocol, each selected source contributes 3--5 representative module skills.

\paragraph{Verification through isolated tests and recovery.} Every module skill receives an assertion-backed test or smoke check. After the modules pass in isolation, \disco evaluates whether their composition can recover a bounded result or mechanism from the source paper. The recovery stage may use the paper, generated module documents and skills, and approved datasets or runtime resources, but it does not read the original implementation repository. This source boundary tests whether the generated skills carry the operational knowledge rather than merely pointing back to the source code. Recovery failures are analyzed at the module level and trigger focused revisions to interfaces, procedures, configuration, dependencies, or evaluation logic.

\paragraph{PaperBench pool and downstream running.} After verification, each target paper receives a pool assembled from the validated skills of its selected related-work sources. The research agent autonomously selects and composes skills from this pool during reproduction. The target-conditioned selection keeps the pool relevant to the current problem, while source anchoring keeps each constituent skill reusable beyond that target. The no-skill condition uses the same backbone, harness, and running budget without access to the constructed pool, so the reported comparison measures the downstream effect of the paper-derived operating context rather than replaying a fixed target solution.

\subsubsection{FrontierCS}
\label{app:frontier-skill-construction}

FrontierCS anchors distillation on the benchmark as a whole rather than on individual tasks. This differs from the per-competition graphs used for MLE-bench and the per-target source-skill pools assembled for PaperBench. We build a single recovery-oriented skill graph shared across the Agent Track, covering exact-solution, heuristic, interactive, and online settings. Throughout construction, we exclude text from individual problems, solution fragments, implementations, task identifiers, and tuned constants. Construction and refinement are completed before evaluation, after which the graph is frozen for all reported runs.

\paragraph{Grounding starts from expert problem-solving guidance.} The initial evidence comprises two human-authored sources. One catalogs common algorithms and data structures together with their applicability conditions and rejection criteria. The other provides quantified guidance for heuristic search, including budget allocation, representation, incremental evaluation, neighborhood design, optimizer selection, calibration, and deadline management. We supplement them with problem-solving guidance adapted from the original workspace prompt, emphasizing reading the full contract, preserving a simple correct fallback, using brute force and randomized differential testing, probing boundary and resource limits, and treating graded submissions as diagnostic feedback.

\paragraph{Construction organizes recovery knowledge as a routed graph.} The entry skill acts as a recovery router and is activated only after a focused initial attempt exposes a concrete failure or leaves the solution incomplete. It reconstructs a compact recovery snapshot, identifies the earliest implicated layer, and routes the agent to one of eight progressively disclosed modules. These modules cover modeling and method selection, implementation, checker and evaluator construction, validation, interactive inference, reactive online decisions, testlib judging, and plateau escape. After repeated references to the same ordered pair are collapsed, the entry skill and modules form the shared skill graph $\mathcal{G}$ with nine nodes and 42 distinct directed links. Across the graph, the agent distinguishes a guaranteed-valid \emph{fallback}, the best-validated \emph{champion}, and experimental \emph{challengers}. A weaker or invalid challenger cannot replace the champion.

\paragraph{Paired trials drive iterative verification.} On selected development problems, each candidate graph is evaluated through matched agent trials. At refinement round \(t\), runs using the current graph \(\mathcal{G}^{(t)}\) are compared with both a fixed no-skill baseline and runs using the preceding version \(\mathcal{G}^{(t-1)}\), under the same backbone, harness, task interface, and scoring endpoint. The former measures the accumulated effect of the graph, while the latter isolates the effect of the latest revision. We inspect score changes together with the paired trajectories, including the chosen formulation and method, data structures, local tests, submissions, sanitized feedback, and final artifact. Repairs are localized to the implicated nodes or links, and a revision is retained only when follow-up trials reproduce the intended behavioral change or provide independent evidence that the revised guidance corrects the diagnosed failure; otherwise, it is excluded from the frozen graph.

\paragraph{Refinement retains recurrent mechanisms.} Positive cases repeatedly preserved a legal champion, changed the implicated problem formulation or search structure, and compared materially different challengers. Negative cases exposed unnecessary evaluator construction, repeated tuning within one method family, and premature delegation. We retain only recurrent, causally supported revisions to activation timing, failure-layer routing, evaluator evidence gates, champion preservation, and plateau escape, guided by quantified criteria and independent critique. Together, these revisions transformed the initial monolithic workflow into the final routed graph.

\subsubsection{PassNet}
\label{app:passnet}

PassNet uses task-oriented distillation. Each benchmark instance is treated as a task anchor \(z=\tau\): the run first determines what the instance requires, then identifies capabilities that the current agent does not reliably provide, gathers evidence from task executions and evaluator feedback, and distills useful procedures into a skill graph. The analyses are performed at the level of individual task instances, while evidence from multiple instances is pooled when revising the shared graph. We keep the backbone and harness fixed throughout construction.

\paragraph{Capability scoping and candidate grounding.}
For a PassNet instance, the relevant capabilities can include inspecting an FX graph, determining whether a pattern is likely to match, selecting regions that may benefit from fusion, writing a Triton replacement, validating semantic equivalence, and interpreting evaluator or match failures. We use an initial attempt to determine which capabilities are already available to the agent and which gaps prevent a useful optimization. Before collecting task trajectories, we provide provisional candidate procedures based on the agent's existing knowledge. These procedures are treated as hypotheses about useful operational knowledge rather than accepted skills. We then run the agent on 50 training-set instances under matched conditions, with and without these procedures, and use the resulting paired trajectories to revise the candidates and assemble the first version of the PassNet skill graph.

\paragraph{Scale-up refinement over the training set.}
The PassNet training split contains more than 4k instances. Solving all of them in full would take more than 20 days and incur substantial API cost, so we use a screening-and-dispatch workflow to search for additional task instances that may reveal useful optimization or failure patterns. A CPU pass first selects screened candidates using a lightweight heuristic for potential improvement over eager execution. We sample-check the screened candidates, dispatch them to sub-agents in batches, and ask the sub-agents to summarize recurring optimization patterns and failure modes observed in their trajectories. We use these summaries, together with the underlying task outcomes, as additional evidence for revising the skill graph. One pass over the training set takes approximately one day and supports the second graph version.
\begin{table}[ht]
\centering
\small
\setlength{\tabcolsep}{6pt}
\caption{PassNet construction-stage results on 50 sampled training tasks with Claude Code harness and DeepSeek v4 pro backbone to economize.}
\label{tab:passnet-construction}
\begin{tabular}{lccc}
\toprule
\textbf{Metric} & \textbf{Baseline} & \textbf{First version skill} & \textbf{Second version skill} \\
\midrule
Arithmetic mean & 0.9566 & 0.7474 & 0.7492 \\
Geometric mean & 0.5352 & 0.6840 & 0.6941 \\
Median & 0.7340 & 0.7818 & 0.7941 \\
Match failures & 10/50 (20\%) & 2/50 (4\%) & 1/50 (2\%) \\
Average solve time & 40 min & 22 min & 22 min \\
\bottomrule
\end{tabular}
\end{table}
\paragraph{Paired verification and failure-driven refinement.}
We evaluate each graph revision on the same sampled tasks with the same backbone, harness, task interface, and scoring endpoint. We compare the no-skill baseline with each graph version and inspect the paired trajectories alongside the aggregate metrics. Table~\ref{tab:passnet-construction} shows that across the two graph revisions, the geometric mean increases from 0.5352 for the no-skill baseline to 0.6941 for the second-version graph, the median increases from 0.7340 to 0.7941, match failures decrease from 10/50 to 1/50, and average solve time decreases from 40 minutes to 22 minutes. The arithmetic mean is more sensitive to outliers in this sample: two instances receive unusually large scores, 10.63 and 8.18, in one no-skill run, but score 0.99 and 0.10, respectively, in another repeated run. Because these outliers were not reproduced reliably, we examined the corresponding skill-equipped trajectories. This error analysis identified an overly conservative rule rather than a missing implementation template. The skill warned the agent against reimplementing vendor-optimized heavy operators, such as large dense matrix multiplications or convolutions, with custom kernels. This is a useful default in many cases, but it blocked a \texttt{conv3d} case with \texttt{stride == kernel\_size}. In this regime, the computation can reduce to elementwise multiplication, while the generic cuDNN path may still incur layout and gather overhead associated with overlapping convolution. The trajectory showed that the agent recognized this possibility but abandoned the rewrite because the warning was expressed too absolutely. The case was absent from both the initial sample and the subsequent screening pass, so the earlier evidence did not expose the overly broad rule. We revised the skill to treat the warning as a defeasible prior rather than an absolute ban: vendor implementations should normally be preserved, but graph structure and evaluator evidence may justify an exception.